# Current Advancements on Autonomous Mission Planning and Management Systems: an AUV and UAV perspective

[1]Adham Atyabi, [2]Somaiyeh MahmoudZadeh, [3]Samia Nefti-Meziani

[1] Seattle Children's Research Institute, University of Washington, Washington, United States
[2] Faculty of Information Technology, Monash University, Melbourne, Australia
[3] Center of Autonomous Systems and Robotics, University of Salford, United Kingdom
Adham.atyabi@seattlechildrens.org
somaiyeh.mahmoudzadeh@Monash.edu
s.nefti-meziani@salford.ac.uk

***Abstract*** Advances in hardware technology have enabled more integration of sophisticated software, triggering progresses in development and employment of Unmanned Vehicles (UVs), and mitigating restraints for onboard intelligence. As a result, UVs can now take part in more complex mission where continuous transformation in environmental condition calls for higher level of situational responsiveness. This paper serves as an introduction to UVs mission planning and management systems aiming to highlight some of the recent developments in the field of autonomous underwater and aerial vehicles in addition to stressing some possible future directions and discussing the learned lessons. A comprehensive survey over autonomy assessment of UVs, and different aspects of autonomy such as situation awareness, cognition, and decision-making has been provided in this study. The paper separately explains the humanoid and autonomous system's performance and highlights the role and impact of a human in UVs operations.

***Index Terms*** *— Cognition, Mission Management, Autonomy, Contingency Management, Situational Awareness*

## 1   Introduction

Increased level of autonomy has resulted in reduction of the human role for supervision. Analyzing encircling situation is the most crucial part of autonomous adaptation. Since there are many unknown and constantly changing factors in the real environment, momentary adjustment to the consistently alternating circumstances is highly required for addressing autonomy. To respond properly to changing environment, an utterly self-ruling vehicle ought to have the capacity to realize/comprehend its particular position and the surrounding environment. However, these vehicles extremely rely on human involvement to resolve entangled missions that cannot be precisely characterized in advance, which restricts their applications and accuracy. Reducing dependence on human supervision can be achieved by improving level of autonomy. Over the previous decades, autonomy and mission planning have been extensively researched on different structures and diverse conditions; nevertheless, aiming at robust mission planning in extreme conditions, here we provide exhaustive study of UVs autonomy as well as its related difficulties in internal and external situation awareness. In the following discussion, different difficulties in the scope of AUVs and UAVs will be discussed.

### 1.1   Challenges that Autonomous Underwater Vehicles (AUVs) Face

Autonomous robotic platforms arouse more interest in recent decade and broadly conducted in maritime approaches to achieve routine and permanent access to the underwater environment. AUVs have exposed their capabilities in cost-effective deep underwater explorations, far beyond what humans are capable of reaching. Hence, nowadays they become the best option for performing autonomous navigations and missions. AUVs are largely employed to undertake various missions such as inspection and survey [1], scientific underwater explorations [2], sampling and monitoring coastal areas [3], measuring turbulence and obtain related data [4], offshore installations and mining industries [5], and military missions [6]. Generally, they are capable of spending long periods of time carrying out underwater missions at lower costs comparing to manned vessels [7]. Autonomous operation of an AUV in a vast, unfamiliar and dynamic underwater environment is a complicated process, specifically when the AUV is obligated to react promptly to the environmental changes [8]. An AUV, thus, needs to ensure safe deployment at all stages of a mission, as its failure is inadmissible due to expensive maintenance. On the other hand, diversity of underwater scenarios and missions necessitates the requisition for robust decision making based on proper awareness of the situation. An overview of the possible environmental situations has a significant impact in enhancing vehicle's overall autonomy. There is always a strict time restriction for AUV's operations especially when they are demanded to carry out highly complex tasks. Currently, AUVs have restricted energy supplier and bounded endurance; hence, they have to efficiently manage their resources to perform efficient persistent deployment in more extended missions without human involvement. Therefore, a human-like intelligence is an essential requirement for advancing the vehicle's autonomy to trade-off within tasks importance and the constraints while adapting the terrain changes during a mission. Having a robust mission-motion planning and task management strategies is preliminary for autonomous operation of such self-controlled systems [9]. However, their operations still remain restricted to very particular tasks with little real autonomy due to reasons which is going to be discussed in Section 2.

## 1.2 Challenges that Unmanned Aerial Vehicles (UAVs) Face

The use of UAVs over alternatives offers several advantages including cost reduction, improvement in flight performance, and expend-ability [10]. The UAVs are generally divided into two categories of remotely piloted and autonomous vehicles. Veres et al. [11] outlined five essential features that make a system autonomous: *i*) efficient energy source, *ii*) structural hardware, *iii*) computing hardware, *iv*) sensors and actuators, and *v*) autonomous software. The autonomous systems share the first four features with the remotely operated vehicles.

Recent technological advancements in the field of UAVs have pushed the boundaries towards autonomous structures that require less and less human supervision and carry a higher level of on-board intelligence. However, 'intelligence' is a difficult concept to grasp, formalize, and measure. Meystel and Albus [12] used the following definition: "*Intelligence is the ability of a system to act appropriately in an uncertain environment, where an appropriate action is that which increases the probability of success, and success is the achievement of behavioral sub-goals that support the system's ultimate goal.*" In an ideal scenario with UAVs, the autonomous vehicles should not rely on human inputs or updates, and should only rely on the information gathered from their on-board sensors. It is necessary to have common grounds for assessing and comparing the level of autonomy achieved by several studies. This paper aims to present some of the recent developments in the autonomous mission planning and management systems in UV studies with a focus on AUV and UAV.

## 2 How to Assess the Level of Autonomy?

One of the common confusions in autonomous systems is the difference between autonomy and intelligence. Clough [13] explained this difference stating: "*They are not the same. Many stupid things are quite autonomous (bacteria) and many very smart things are not (my 3 year old daughter seemingly most of the time)*".

One can describe Intelligence as the ability to discover knowledge and use it to do something. On the other hand, in an agent, autonomy can be defined as the agent's ability to generate its own purposes without requiring outside instructions. As Clough [13] discussed, what is needed to be known is how well a UAV do the associated tasks to reach goals when no operator is around to oversee the mission's moment to-moment operation. Therefore, one might say that it is not important how intelligent the system is but if it is capable of performing the assigned jobs. Hence, a vehicle's autonomy can be investigated in many ways and different levels according to the expectation of vehicle's performance.

Highly complex approaches with high levels of autonomy still rely on expert's observation and control, so the platform always requires the involvement of an expert's knowledge and suggestions. Therefore, to make AUVs genuinely reliable and autonomous, more advanced Situational Awareness is needed [14]. For a successful mission, the autonomous system should continuously monitor its resources and reorder the tasks to maximize mission performance during the operation. Higher autonomy can improve UVs' operation in the following areas:

- Energy consumption: optimize the power sources and vehicle's power consumption to enhance its capabilities in long endurance missions.
- Risk-free optimal navigation: navigating and positioning correctly with minimum position calibration error.
- Robust decision making: autonomy in this discipline can be defined as the capacity of sensing, interpreting and acting upon unexpected or unknown changes of environment. An autonomous vehicle should be capable of making efficient decisions when facing different challenging situations.

Autonomy in decision-making and navigation are connected in various standpoints. A UV must be highly intelligent to catch the operators trust by its survivability, recovering from faults (if applicable), and carrying out the mission efficiently [8]. One essential concern for autonomy and efficient mission planning is the robustness of a vehicle's path/trajectory planning in order to cope with strong environmental variability. For instance, in the case of underwater operations, AUVs' communication is yet restricted with high latency and excessively low bandwidth. Therefore, AUVs have very little chance for communicating with the operators and exchange data during the mission. Moreover, proper estimation of the underwater behavior, beyond the sensor coverage, is impractical and inaccurate. Present technology is only capable of forecasting limited components of the ocean variability. This limited knowledge about the future state of the environment reduces AUV's autonomy, safety, and its robustness. Thus, the system should dynamically track the available time, the lost time, and environmental changes; then it should be able to manage its resources by adjusting its parameters according to the new environment state, which helps to maintain its operation at near optimal level. Given the importance of path-planning in overall success of UV missions, different strategies have been introduced. The well-known direct method of optimal control theory, called Inverse Dynamics in the Virtual Domain (IDVD) method, is used to develop and test an onboard real-time trajectory generator for AUVs' deployment and docking procedure [15, 16]. A real-time Differential Evolution (DE) based motion planner is designed by MahmoudZadeh et al. [17] to provide a time/battery efficient operation for a single AUV in a dynamic time-varying ocean environment. Later on, another elaborated evolution-based online path planner/re-planner is introduced for AUV rendezvous path planning in an uncertain underwater terrain to ensure AUV's safe deployment and secure docking [18]. The authors assessed the feasibility of four bio-inspired algorithms in their effort towards developing a robust planner capable of coping with ocean current variability and terrain uncertainties. In their study, the effectiveness of the vehicle's SA and its ability to autonomously react to different scenarios is assessed aiming to assure AUV's safe deployment and mission success.

Higher autonomy can improve UV's capabilities in handling mission objectives, where these abilities are substantially influenced by navigation system performance [19]. The trajectory planning techniques are specifically designed to deal with quality of vehicle's motion while facing deviation from expected environmental characteristics. A fully autonomous vehicle should have the ability to consider its position as well as, its environment in order to react appropriately to unexpected events. In this section, four main methodologies for assessing the level of autonomy are presented.

## 2.1 Mobility, Acquisition, and Protection (MAP)

MAP has been developed by Mark Tilden at Los Alamos national Lab to assess the autonomy level of robots [20]. This method uses mobility (the capability of utilizing movements in the environment with $M_0$ reflecting no motion ability and $M_x$ representing the ability to move in $x$ dimensions with maximum $x$ of 5); acquisition (the ability to extract, store, and utilize energy with $A_0$ indicating zero energy consumption or delivery and $A_x$ representing capability to use planned tactics to gather, store, and utilize energy with maximum $x$ of 5); and protection (the self-defend-ability with $P_1$ representing flight/hide behaviors and $P_x$ indicating higher level of fight/flight capabilities against hostile stimuli with maximum $x$ of 5). The proposed autonomy classification by Tilden makes use of a radar view interface (which makes it more desirable for army use), but it has limitations when multi-UAV scenarios are required due to *i*) the acquired fixation on some of the vectors in the radar view caused by the assessment of multiple UAVs at the same time, *ii*) its inability to address operational characteristics of UAVs, *iii*) its inability to address UAV's interactions, and *iv*) weak discrimination between various levels of autonomy.

## 2.2 Draper Three Dimensional Intelligence Space

Stark's metric, developed at the Draper Laboratory [21], aims to measure both intelligence and autonomy, which contains useful features such as using three metrics that allows the outcome to be presented by a radar view and also considering the operational issues in the assessment. Although the resulting metric is capable of addressing the multi-UAV autonomous control system, it has some other limitations and drawbacks including *i*) the multi-UAV scenario will max-out the radar view due to the poor resolution of the metric; *ii*) the metric uses task planning in one of the radar view axes while task planning is not necessarily an indication/pre-requisition of autonomy; *iii*) SA is measured and quantified based on the number of sensors in the system and the utilized fusion method rather than clarifying if the system is capable of understanding the on-going events happening around it that cause the changes in the output of the sensors.

## 2.3 Autonomous Control Level Chart (ACL)

Clough [15] introduced a new classification/taxonomy in order to assess the autonomy level of UAVs. The proposed classification method contains 9 levels of autonomy in which radio-controlled drones are placed at the lowest level (level 0), with each higher level achieved through the addition of an extra set of functionalities to the previous level, and the highest level (level 9) hosting the fully autonomous vehicles. These extra functionalities and capabilities that allow a UAV to reach a higher level of autonomy are:

- Autonomously executing a pre-planned mission (level 1).
- Autonomously switching between a set of pre-generated mission plans concerning the UAVs' SA (level 2).
- Adding sophisticated capabilities such as fault mitigation and contingency management in single UAV scenarios (level 3 to 5).
- Coordination and cooperation capabilities in multi-UAV scenarios (level 6 to 9).

## 2.4 Sheridan Scale for Autonomy

Tom Sheridan [22] introduced a scale metric known as Sheridan's scale with 10 levels of autonomy in which the lowest level represents an entity that is entirely controlled by a human (tele-operated), and the highest level represents an entity that is completely autonomous and does not require human input or approval. The various levels of autonomy presented by Sheridan are as follows:

- Fully controlled by operator.
- Computer offers action alternatives.
- Computer narrows down the choices.
- Computer suggests a single action.

- Computer executes that action upon the operator's approval.
- Operator has limited time to veto the computer's decision before its automatic execution.
- Computer executes commands automatically, and then the operator would be informed.
- Computer informs the operator after automatic execution if the operator requests it.
- Computer informs the operator after automatic execution only if it decides to.
- Fully controlled by the computer (ignoring the operator).

In addition to these four methodologies, Veres et al. [11] considered a simple 3 level of autonomy. In their model, the first level included vehicles that are capable to autonomously tracking their self-generated trajectories to specific way-points. The second level contained vehicles that are capable of navigating towards self-defined intermediate way-points while the human supervision is limited to identification and presentation of the global targets. Ultimately, the third level included vehicles that have all capabilities of previous level vehicles in addition to the ability to receive mission goals from human operators and perform extensive decision-making and on-board mission-related knowledge analysis. The proposed metric is general, and it is not adequate for proper classification of various existing autonomous systems.

### 2.5 Case Study: Comparison of Various Measures of Autonomy

Generally, autonomy for UVs can be considered in three primary levels of fully autonomous, semi-autonomous and fully human operated systems. In order to better understand the differences between various measures of autonomy mentioned earlier, a hypothetical case study is being presented in this section.

To start, we consider a simplistic scenario where an AUV should go from point $a$ to $d$ and visit points $b$ and $c$, deal with some harsh water current and some obstacles on its way. The performance of the vehicle is discussed for three different levels of autonomy:

***Fully autonomous operations:*** the vehicle is adjusted and stabilized by feedback control and can autonomously route its trajectory to assigned waypoints. At this level, the vehicle is capable of locating its position on the map and follows the defined path, without operator's interaction. Given a candidate sequence of waypoints and prior environmental information, the path planner provides trajectories to safely guide the vehicle to move through the waypoints until it reaches to the destination. After visiting each waypoint, the time spent to reach that waypoint is compared with the expected time. If the vehicle is behind its predetermined schedule, re-planning, task rescheduling, or mission objective reprioritization will be performed to maximize the chance of success in delivery of mission considering the remaining tasks, mission objectives, and available resources. Such AUV takes advantage of several routines that provide capabilities such as real-time path planning, real-time SA, real-time contingency management, and task scheduling.

***Semi-autonomous operations:*** In this example, such vehicle has operator oversite with SA and autonomous path planning and is usually incapable of performing contingency management, path re-planning or task re-scheduling. In this example, the vehicle's navigation system has capability of indicating intermediate waypoints, which allows the human operator to define main targets without giving details. The system has also capability of recognizing the situation, planning a trajectory between points, autonomous departure, return and collision avoidance; however, the operator decides how to re-arrange tasks (visiting waypoints of $a, b, c, d$) to compensate the lost time and re-plan a mission scenario.

***Fully human operated:*** such a system is fully controlled and monitored by a human operator; decisions are taken by the human expert considering inputs, some previous experiences, and primitive knowledge. The operator starts to move the vehicle from point $a$; plans to visit point $b$ then $c$, and reach point $d$ before running out of battery. In the midway of moving from $a$ to $b$, an unforeseen situation arises (such as appearing a buoyant obstacle or facing sudden water turbulent). The situation severity is recognized relying on operator's intelligence. Then the operator decides how to direct the vehicle to cope with the situation and safely pilot the vehicle toward the point $b$; then, decides to skip visiting point $c$ for on-time arriving in $d$ and compensating the wasted time in the previous stage. The identical aspect of these systems is that they are relying on human SA and decision-making to deal with any unexpected circumstance in a continuously changing environment. The human operator specifies high-level mission scenarios and the vehicle holds extensive mission-related information. However, human operators occasionally have limitation associated to low level of attention, tiredness, miss precaution, and vagueness or improper cognitive capabilities.

In order to better understand, the utility of the discussed measures of autonomy, three of these methods, e.g., MAP, Draper and ACL, are utilized to assess the level of autonomy in our fully autonomous, semi-autonomous and human operated AUVs. The results are presented in Fig.1. It should be noted that the presented graphs in Fig.1 only demonstrated the portion of the scale that is occupied by our three hypothetical AUVs rather than reflecting the entire range of the measures. For example, while in Draper (1b) the measure of autonomy expands to a range of 0 to 9, we are only presenting 0-4 levels aiming to magnify the subfigure for better viewing. In addition, while the original demonstration of these three methods included a radar type of illustration which incorporated rings for each level of autonomy, we have decided to eliminate those extra rings in the cognizance of better viewing of the outcome and using a simplistic triangle representation.

From the results presented in Fig 1, it is clear that neither of these measures can truly highlight the differences between these 3 case scenarios and their measures are subject to interpretation. Draper (Fig 1b) in particular can show no differentiation between how automated the system is since all the measures and markers are based on functionality rather than where the functionality is coming from. In such measure, an AUV with robust and optimum communications that can get guidance from human operator (fully or partially) is going to have the same functionality and capabilities that a fully autonomous AUV poses. This issue is observed in Fig 1 with the 3 AUVs (autonomy scenarios) being drawn on top of each other resulting in a unified triangle (illustrated by red lines indicating that the 3<sup>rd</sup> scenario was drawn last). MAP (Fig 1a) is showing clear differences between the three scenarios however, only the best possible conditions are mapped here. For example, in human operated AUV scenario (red lines), it is assumed that constant and reliable communications exist so the operator can have optimal control over the actions of the vehicle while having reliable and clear sensory readings in AUV missions are rarely possible specially in complex missions where the AUV has to perform deep dives, beyond the feasible length of wired communication and where the wireless communication is also not possible or it is unreliable. It is noteworthy that MAP results indicate highest functionality for human operated AUV while in practice, the range of missions that a fully autonomous AUV can cover supersedes human operated version. Among the three assessment methods, ACL (Fig 1c) shows the best differentiations however, this measure is also subject to interpretation. For example, although we allocated high scores to our fully autonomous AUV in this measure (blue line), realistically, this level of scores in this measure is preserved for multi-UV scenarios despite the fact that in our scenario our AUV poses all functionalities encapsulated in such level. Examples of these functionalities and their associated scores are 1) detection and tracking of other vehicles within environment (Perception/SA=9), 2) continuous mission/trajectory evaluation and replaning (Analysis/Decision making=8), and 3) onboard collision avoidance (communication/Cooperation=6) (see [13, 15] for more details). In contrast, the highest single UV scores relevant to capabilities of our fully autonomous AUV for these categories are 4, 5 and 5 respectively which project our fully autonomous AUV (blue line) very close to the semi-automated AUV (orange line).

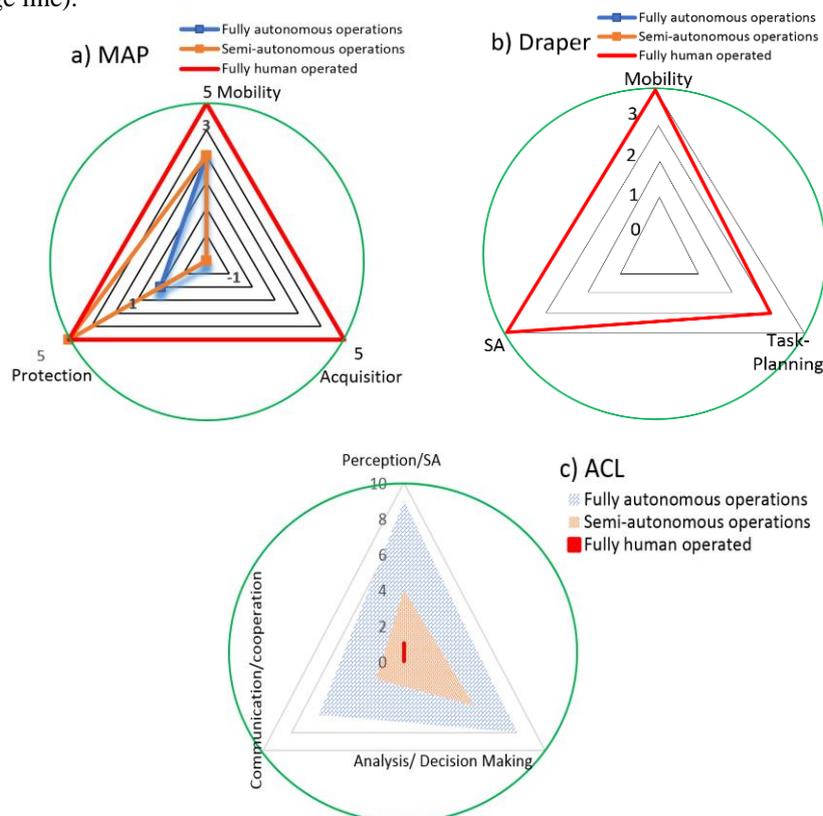

**Fig.1.** Assessing autonomy level on fully autonomous (blue), semi-automated (orang) and human operated (red) AUV scenarios with *a)* MAP, *b)* DRAPER and *c)* ACL (see [13] for detail description of measures).

This unreliable nature of existing mechanisms for measuring the levels of autonomy in general and inability of such measures to capture multi-facet nature of autonomous systems, mission scenarios, and the associated human-machine collaboration in particular is clearly reflected in 2012 final report of the task force to DoD [23].

*"The Task Force reviewed many of the DoD-funded studies on "levels of autonomy" and concluded that they are not particularly helpful to the autonomy design process. These studies attempt to aid the development process by defining taxonomies and grouping functions needed for generalized scenarios. They are counter-productive because they focus too much attention on the computer rather than on the collaboration between the computer and its operator/supervisor to achieve the desired capabilities and effects. Further, these taxonomies imply that there are discrete levels of intelligence for autonomous systems, and that classes of vehicle systems can be designed to operate at a specific level for the entire mission. These taxonomies are misleading both from a cognitive science perspective and from observations of actual practice. [...] In practice, treating "levels of autonomy" as a developmental roadmap has created a focus on machines, rather than on the human-machine system. This has led to designs that provide specific functions rather than overall resilient capability.*

*The Task Force recommends that the DoD abandon the use of '**levels of autonomy**' and replace them with an autonomous systems reference framework that explicitly:*

– *Focuses design decisions on the explicit allocation of cognitive functions and responsibilities between the human and computer to achieve specific capabilities,*

- *Recognizes that these allocations may vary by mission phase as well as echelon and*
- *Makes the high-level system trades inherent in the design of autonomous capabilities visible"*

Vehicle's awareness of environment must constantly be updated due to persistent change and transformation of the situations. This forces the human operator to adapt many cognitive strategies for maintaining SA in highly dynamic environments. Experience can also result in a particular level of automaticity in mental processing of situations that lead humans' automatic reaction in facing a similar situation to their previous experience. The pattern-recognition and automatic action-selection can be considered as primary steps to mimic human cognitive strategy achieving fully autonomous self-controlled functions, which can also improve SA for more demanding tasks. Humans have the general cognitive capability to concentrate on the relevant events and ignore the others.

For instance, consider a fish as a semi-intelligent agent. A fish moves to different positions and avoids colliding obstacles, while it is often unaware of how its body moves to avoid obstacles, but in contrast, it can concentrate on where it wants to go. Automaticity is advantageous to SA because it requires very little conscious attention to process and extricates mental resources for other tasks. Before discussing various components associated with autonomy, mission planning, and mission management systems, it is necessary to discuss the differences between autonomy and automation. This discussion helps to gain a better understanding of the expectation from an autonomous vehicle.

## 3 Autonomy vs. Automation

There is a big difference between two concepts of automatic and autonomous operations. In automatic systems, the machine exactly follows the programmed commands, and it has no choice for making decisions. In contrast, the autonomous systems have the capability of recognizing different situations and making a decision accordingly without human interaction. Hence, developing more intelligent autonomous systems with the ability to run autonomously, reconfiguring based on varying situations, and efficient mission planning according to new circumstances, is one of the significant fields of interest for many system designers [24, 25].

Automatic systems play an essential role in the modern flight control systems and have potential in terms of increasing efficiency, comfort, and safety of the motion, but they should not be mistaken with autonomous systems. Stenger et al. [26] provided the following explanation to distinguish between the two categories: "*an automatic system is designed to fulfill a pre-programmed task. It cannot place its actions into the context of its environment and decide between different options. An autonomous system, on the other hand, can select amongst multiple possible action sequences in order to achieve its goals. The decision which action to choose is based on the current knowledge, that is, the current internal and external situation together with internally defined criteria and rules.*"

Autonomous systems can also be seen and explained in the context of cognitive systems. Vernon [27] explained the relationship between cognition and autonomy as "*One position is that cognition is the process by which an autonomous self-governing agent acts effectively in the world in which it is embedded. As such, the dual purpose of cognition is to increase the agent's repertoire of effective actions and its power to anticipate the need for future actions and their outcomes*".

Following these principles, an autonomic system must be capable of sensing, monitoring, and understanding of all its operational context and probable environmental events. It should adapt to the variable uncertain environments with minimum human involvement. This process is strictly tightened to the definition of SA proposed by Endsley [14]. An autonomous vehicle should also be aware of its internal state and capabilities to assess whether current mission goal is reachable or find an alternative solution.

## 4 Mission planning and Management Systems and Their Components

For any of scientific, mine or military applications of UV, a sequence of specific tasks are predefined and characterized in advance and they are fed to the vehicle in series of command formats. A vehicle should carry out complex tasks in pre-specified time intervals and has to manage its resources effectively to perform persistent deployment in longer missions. Limitations on battery capacity and restricted endurance arose the necessity of managing resources and mission time. A typical autonomous mission management system constitutes of components such as mission planning (task/resource allocation, motion planning), mission execution (navigation, task execution, intelligent decision making), mission monitoring (mission progress evaluation, situational awareness, contingency and anomaly detection), and mission re-planning (re-tasking, resource reallocation, re-routing).

MahmoudZadeh [28] defined mission planning as a joint problem of discrete natured task priority assignment and continuous natured vehicle routing toward the destination in graph-like terrain. They mapped the distributed tasks in a specific operation area by a network, which took the complication of the vehicle routing in a graph-like terrain into account. The task assignment in this context considered as a sub-process in routing and the mission planning framework tend to maximize mission productivity in terms of battery lifetime management and optimality (order and priority) of the selected tasks considering different characteristics such as tasks priority value, risk percentage, and absolute completion time, along with vehicle's availabilities and capabilities. A typical architecture of a mission management system is shown in Fig.2. Although the presented architecture assumes human in the loop with an interactive and corrective role. In an autonomous architecture, the human operator role is more in the sense of observatory with minimum interaction/correction. The presented architecture in Fig.2 assumes a level of on-board intelligent and independent decision-making ability for individual/swarm UAV(s).

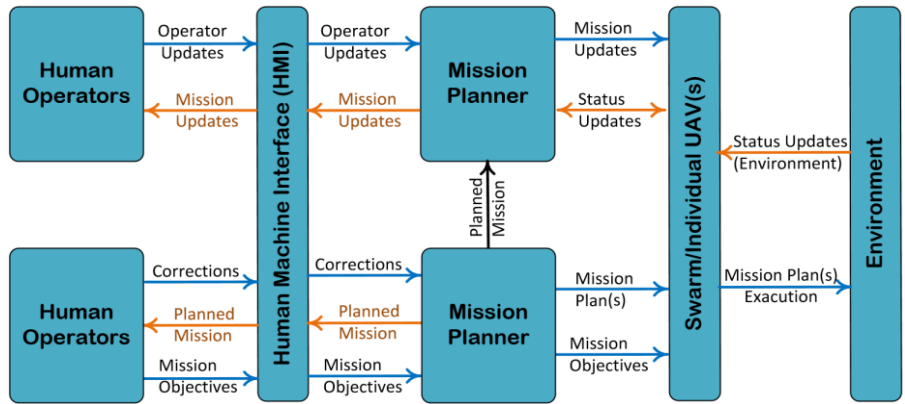

**Fig. 2.** High level architecture of a typical mission planning and management system with human operators in the loop.

An alternative approach is to transfer the decision making and task autonomy plans of UAV(s) to a ground control module where the sensory data of vehicle(s), their perception of the environment, and their tasks/roles/objectives are received through the networking communications. Afterward, the mission status is updated, and new actions are transmitted to UAV(s). In this architecture, mission manager is tasked to map the mission progress to the mission plan, assess mission health, assess mission objectives' achieve-ability, detect contingencies, and produce new plans if necessary. Following sections discuss some of the existing mission management systems and their contributions for UAVs and AUVs.

### 4.1 What is Situation Awareness (SA)?

The capability of understanding and dealing with extremely dynamic and intricate environments is defined as SA, and it is an essential requirement for systems that are obligated to react to a dynamic and variant environment. SA is an idea to illustrate the process of sensing, comprehending and operating in any environment. Improving the SA for UVs can promote their capabilities from full human control to entirely autonomous control [3]. Unforeseen and unpredictable conditions can enforce the mission to abort and sometimes might even cause the loss of the vehicle, as occurred for Autosub2 that was lost under the Fimbul ice shell in the Antarctic [29].

The current state of a vehicle and vast incoming data flow from surroundings must be integrated and considered in order to achieve a proper mental model of the situation. This integrated picture generates the central organizing schema from which all actions and decision-making take place. SA thus includes three aspects of observation and perception of the environment; understanding and comprehension of the events; and projection of the future possibilities (see Fig. 3).

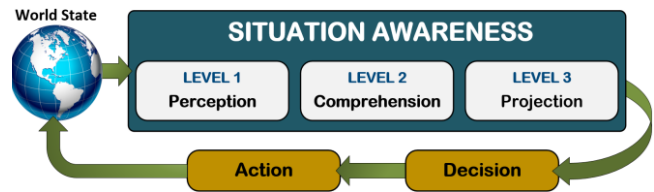

**Fig.3.** Situation awareness and decision making

Systems perception of environment can promote its capability of dealing with unstructured and unforeseen events. Therefore, it is required to process the objects and their characteristics simultaneously through pre-attentive sensory stores. Newly perceived information is combined with existing knowledge in working memory to create a new or updated mental picture of the changing situation. These pieces of information employed to create projections of what may happen in the future. These projections, in turn, help the UV decide what actions (real-time) to take as a result. SA, thus, is one of the most critical required factors for facilitating the next generation of AUVs and UAVs.

Patron et al. [30] demonstrated the process of semantic world model framework in a hierarchical representation of the sensory extracted knowledge. This mechanism could promote the global (system level) and local (agent level) SA. Furthermore, the mission could be parameterized and executed based on the available knowledge from platform capabilities applying a declarative goal-based mission planning approach. The combination of these elements facilitates the vehicle for dynamic mission adaptation to environmental parameters and dealing with the internal issues. Later on, Patron et al. [31] combined the advantages of the knowledge-based framework and the goal-based mission planning to provide interoperability of embedded intelligent agents for on-board decision making. The research represented the progress of independent service-oriented agents' coordination with the adaptive goal-based planner taking the advantages of the situation-aware knowledge-based framework to cover the mission requirements. The proposed mechanism promoted interoperability of SA and the embedded service-oriented agents in autonomous platforms.

Patron et al. [32] utilized composition of hierarchical ontological representation of knowledge and adaptive mission plan repair techniques in his later research, to implement a system capable of adapting mission plans autonomously in the face of events during a mission. They presented a semantic-based framework to provide the core architecture of knowledge representation for service-oriented agents in Unmanned Underwater Vehicles (UUVs) in order to improve the vehicles SA. The system used a goal-oriented approach in which the mission was described in terms of 'what to do' instead of a 'how to do.' The mission was parameterized and executed based on the available knowledge and vehicle's capabilities. Patron was the first researcher who applied a goal-based planning approach for adaptation to the underwater mission and to maintain platform's operability.

Furthermore, Chai & Du, [33] designed another SA-base framework, including event extraction and correlation, force structure recognition and intent inference and prediction, where the expert knowledge system was constructed for event-rule extraction. Miguelanez et al. [34] also proposed a semantic world model framework for the hierarchical distributed representation of knowledge in autonomous underwater systems to increase interoperability, independence of operation, and enhancing SA of the embedded service-oriented agents for autonomous platforms. Jones et al. [35] described an efficient approach to provide an actionable model of SA using Fuzzy Cognitive Maps (FCM) that encompassed all three levels of SA (e.g., perception, comprehension, and projection). The presented cognitive model (SA-FCM), is generated directly from the goals, decisions, and essential information associated with proper decision-making in a particular domain, which is known as a computational naturalistic decision-making model. They attempted to develop a model that replicates human cognition as it relates to SA.

## 4.2 What is Cognition?

Cognition is the combined ability to understand and anticipate how the things might possibly behave or evolve now and in the near future, and to consider this in determining the necessary actions to take. 'Cognitivist' and 'Emergent' are the two main categories of cognitive systems.

Cognitivist systems utilize symbolic representation and processing of information whereas emergent systems are based on the self-organizing principles mainly embraced by connectionist, dynamic, and enactive systems. In the context of the behavior-based representation of a system, cognitivist approaches are defined by sets of behaviors dictated by the experts to deal with a-priori known possible states and, in this sense, sensitive to the inaccurate perception of the environmental state. In contrast, emergent models are capable of learning and evolving behaviors to compensate for unforeseen and unpredicted environmental states.

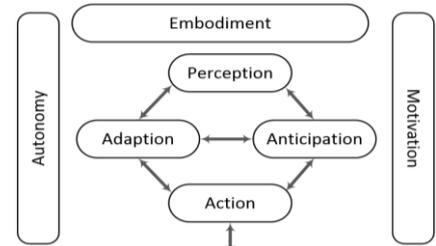

**Fig.4.** Developmental components of cognitive

Vernon et al. [36] explained that while the cognitivism provides an abstract and self-contained model in which physical instantiation of the system does not influence the cognition model, the emergent-based architectures provide a pivotal role for physical instantiation in the cognition model. Figure 4 illustrates the developmental components of a general cognitive system. This development guideline of the cognitive system's architecture indicates that in such a system the actions are influenced by perception (guidance), goals (shaping and directing), and affective motives (triggering). From a need to predict outcomes and to efficiently deal with dynamic events, cognition also provides an ability to construct explanations and imagine unforeseen events. These capabilities can be utilized to expand the agent's repertoire of actions and enhance its ability to efficiently interact with the world around it, all the while maintaining its autonomy. Vernon [27] categorized some of the well-known cognitive-based robotic architectures into three groups:

i. Cognitivist (Soar, Executive Process Interactive Control (EPIC), Adaptive Control of Thought-Rational (ACT-R), and the ICARUS Cognitive Architecture, ADAPT).
ii. Emergent (Autonomous Agent Robotics (AARs), Global workspace, Self-Directed Anticipative Learning (SDAL), Self-Aware Self-Effecting (SASE), DARWIN).
iii. Hybrid (HUMANOID, Cerebus, Cog: theory of Mind, Kismet).

Among the existing cognitive architectures employed in robotic studies, Soar is often utilized for mission management purposes. The Soar system is a rule-based system that utilizes two cycles of production and decision. In the production cycle, those productions that match the declarative memory contents fire which result in alteration of the declarative memory and further firing of other productions. In the decision cycle, a single action would be selected out of all possibilities. Soar uses a sub-goal mechanism whenever an impasse situation occurs. Impasse is the situation in which the actions are ambiguous, or no action is available. In the sub-goal mechanism, the impasse is resolved by setting the new state in a new problem space. This change of state in Soar system is the only form of learning that exists in this architecture [27].

Enactive models are a variation of Emergent systems that improve emergent capabilities by considering cognition as a process whereby essential factors for the continued existence of a cognitive agent are enacted [36, 37]. In enactive models, symbolic representation is not necessary given that no preset information or rule is required, and the system uses an enactive interpretation. Within the concept of cognitive robotic systems, the principal idea of enaction is that the UV develops its understanding of the world around it through its interactions with its environment. Thus, enaction entails that the UV operates autonomously and generates its own models of how the world works. There are five key elements to be considered when dealing with enactive systems:

i. **Autonomy:** The UVs will be endowed with some self-maintaining organizational characteristic akin to living creatures that use their own capacities to manage their interactions with the world, and with themselves, to survive. This means that the system is entirely self-governing and self-regulating. The system take actions that best benefit its own and the swarm's objectives. So the system is not controlled by any outside agency, allowing it to stand apart from the rest of the environment and operate independently. That is not to say that the system is not influenced by the world around it, but rather these influences are brought about through interactions that do not threaten the autonomous operation of the system.

*ii.* **Embodiment:** In enactive cognitive systems, it is necessary for the system to exist in the world physically to directly interact with the environment and objects (inanimate or animate, cognitive or not) and to be influenced by and react to these stimuli. In an enactive cognitive system, different types of embodiment can be considered including Structural Coupling (two-way perturbation of a system and the environment), Historical (history structure coupling), Physical (capability to have forcible action), Organismoid (for humanoid or rat-like robots), and Organismic (autopoietic living systems) embodiments.

*iii.* **Emergence:** This element addresses the way cognition arises within the individual and swarm of UVs. The emergence refers to the set of laws and mechanisms which govern the behavior of the components of the UV(s). In this system, the behaviors (cognitions) arise from the dynamic interplay of defined laws and mechanisms with the existing components of the cognitive system. In other words, the internal dynamics which are essential for maintaining the autonomy of the system emerge from the controlling behaviors. It should be noted that one of the objectives of this proposal is to develop an architecture with a multi-layered hierarchical level of autonomy and intelligence capable of facilitating the emergence and sense-making requirements in a complex cognitive system.

*iv.* **Experience:** This element is the UV's history of interactions with its surrounding world. Such interactions trigger state changes (structurally determined) in the system rather than controlling the system. These changes depend on the system structure regarding the embodiment of the self-organizational principles as the necessary requirement of autonomy.

*v.* **Sense-making:** Sense-making refers to the relationship between the knowledge encapsulated by the UV and the interactions which gave rise to it. The system generates emergent knowledge that captures some regularity or lawfulness. However, the sense it makes is dependent on the way in which it can interact: its own actions and its perceptions of the environment's reaction.

Some researchers emphasized on cognitive-based decision-making methods. Xu, Yang and Zhang [38] designed a Cooperative Mission Management System (CMMS) with a distributed architecture for a team of cooperative agents. The architecture distributes the control sub-systems between the leading agent and the ground control station. The system considers two layers of management in which the CMMS assigns mission objectives to team members without causing conflict with the sole autonomous management system that is on-board and is focused on resource allocation and management/scheduling. Within the system, the ground control unit is responsible for assigning the mission objectives before deployment of the agents, monitoring their status and providing a real-time feed to operators during the mission while the team-leader is responsible for the sole autonomous management and correlating other team members. The proposed CMMS architecture in [38] provides various functionalities for a team of heterogeneous agents such as SA, team forming, mission assignment, resource scheduling and health management.

Hu and Liu [39] proposed a mission task assignment methodology constrained by health management issues, where the robots' capacity to returning to the base after the completion of the mission is taken into account for increasing efficiency of the UAVs in multi-UAV mission scenarios. The proposed methodology divides the team members into groups of decision and member agents with each decision agent being in charge of managing and coordinating its co-team agents. In this system, the member agents in each state of the same team evaluate themselves and their decision agent, and the agent with the fittest evaluation result switches its role with the decision agent of the team.

Boehm et al. [40] presented a two-layer cognitive architecture for resource management. In this hierarchical controlling system, the top-level node contains a cognitive architecture that provides knowledge-based mission management through elements such as: inference-based situation interpretation; goal-driven decision making; search-based planning; and pattern-based task execution. The automation is facilitated by lower level nodes that provide route-planning, data linkages, and a flight control system. The architecture is assessed for flying a fixed wing motor-glider UAV in a real flight.

Another cognitive approach is presented by Stenger, Fernando, and Heni [26] in an autonomous UAV-based mission planning study. The proposed architecture is based on Soar architecture and develops a cognitive agent that provides autonomous capabilities in UAVs within a dynamically changing environment. In this architecture, the higher-level planner develops the overall mission plan (based on a well-known environment assumption), and the cognitive agent architecture (Soar) provides autonomous reactions to unforeseen events and circumstances (based on partially known environment). Authors mentioned that requiring a small amount of pre-programmed task sequences in Soar is advantageous since in Soar's architecture the learning mechanism that is based on interaction with the environment results in evolving actions. The feasibility of the proposed architecture is evaluated in a simulated environment under missions with different stress levels, and the results highlighted the potential of the Soar in addressing the required level of autonomy in UAVs involved in high stressed missions. Russo et al. [41] presented a brief report on the utilization of a fusion-based architecture that takes advantage of cognitive models such as Soar, ACT-R, and Swarming in UAV mission planning problems. The report mainly highlighted the hardware aspect of the architecture, and the results indicated that utilization of specialized hardware in the kernel of the cognitive framework results in achieving significantly better performance in comparison with generally purpose processors.

## 5   The UAV-based Mission Planning and Mission Management Systems

Johnson et al. [42] argued that current UAV's do not satisfy the degree of safety required for operation on civilian and populated airspaces. The study also highlighted aspects such as a vehicle's structure, perception, rapid adaptation, and reflexive response as some of the drawbacks of current UAV technologies resulting in a poor mishap of UAVs in comparison to piloted systems. Although a higher mishap might be acceptable given that there is no loss of human life, it will restrict the employment of technology to non-populated areas. The study tabulated causes of mission failure in UAVs such as emergency procedures (26%), parts quality/suitability (16%), testing (16%), SW configuration control (13%), redundancy of critical systems (10%), design problems (6%) and assembly errors (3%) and argued that it is possible to decrease the occurrences of error by approximately 30% through providing better mission management. Aiming for robust and reusable software architectures that would allow UAVs to monitor their current state, forecast their future states and also address their current problems, Johnson et al. [42] introduced an Autonomous Control Executive (ACE) framework as part of their mission management architecture called TRAC. TRAC is responsible for managing the real-time execution of the mission plane in a step-by-step manner with the following functionalities:

– Issuing mission segment commands,
– Monitoring segment completion,
– Reacting to unexpected events during a mission segment,
– Managing start-up and shutdown,
– Logging mission events and data,

Sullivan et al. [43] reported the development of Intelligent Mission Management (IMM) designed for UAVs with the focus of improving the mission success and reducing the mission risk through providing capabilities that allow UAVs to redirect the flight according to the changes in environmental conditions and goals of the flight (goal-directed autonomy). To this end, an on-board Intelligent Agent Architecture (IAA) and a ground-based Collaborative Decision Environment (CDE) are employed. The CDE system provides SA, collaboration, and decision-making tools for UAV in order to furnish its mission plan, schedule monitoring, direct the payload system, integrate sensing goal to the mission plan, and provide visualization. The system uses an automated planning tool known as MAPGEN (Mixed Initiative Activity Planning Generator) which is a ground-based decision support system that allows human operators to monitor the automatically generated plan and takes necessary steps to keep the mission within the boundary of the available resources. The developed package corresponds to level 6 of Sheridan's scale of autonomy.

Franke et al. [44] discussed the importance of contingency management in mission planning. Contingency management is known as the ability to compensate and recover from unanticipated events and conditions that can affect the plan execution. The study introduces the concept of holistic contingency management that features key elements such as: multi-level assessment; plan-based assessment; capability-based assessment; predictive assessment; and team-based assessment. The study provides a holistic view of contingency management in 2 to 3 structural layers. In one dimension the holistic view included short, medium and long-time spans and in the other dimension, 5 levels of mission management functions are presented including a closed loop response, reflexive/reactive response, plan-dependent monitoring, predictive assessment, and team assessment.

McManus et al. [45] investigated the possibility of increasing on-board intelligence through mission planning and mission piloting. The study noted that other attempts in the field have not considered airspace boundaries and their associated characteristics, as well as to mostly operating within 2D environments while real-world situations are 3D+, concluding that currently developed technologies for UAVs are mostly inefficient for civilian airspace. The proposed package contained modules that satisfy SA, autonomous mission planning, and collision avoidance. The SA module operates by generating a digital representation of the world using triangular meshes in which entities such as terrain, airspace boundaries, adverse weather, other flying objects, tall buildings, and radio frequency zones are identified. The study employed Cub-Space (C-Space) mission planning, which in principle represents the world in 3D cubes that are marked either free or occupied. In this representation, the cubes that overlap with air-space boundaries are marked as occupied. The free cubes are given a weight based on their distance to the goal, and the path is generated based on a set of free cubes that drive the UAV toward the goal with the shortest possible distance. In this approach, the mission planner's task is to identify a set of adjacent free cubes that satisfy the objectives such as: having the shortest distance, require less fuel consumption, and requiring less flying time.

Linegang et al. [46] considered aligning the operator's conceptualization of the mission planning as the most challenging obstacle for the development of intelligent autonomy. Traditionally, the human operators played two roles: specifying mission's goals and constraints; and reviewing, approving, executing, and if necessary overriding the planned actions for satisfying the mission's goals in both off- and online phases. Linegang et al. argued that one source of conflict between the operators and the automation originates from the differences in terms of the considered definitions of errors among them (e.g., not alerting enemy in human operator and reducing fuel consumption in the automation process). The authors suggested the use of interface displays that direct the attention of the operators toward the aspects of the plan that are likely to influence the automation system's functioning and made use of Mission Displays for Autonomous Systems (MiDAS) that uses Cognitive Work Analysis (CWA) in its analysis of the ISR (Intelligence, Surveillance, and Reconnaissance) work domain.

Patron et al. [47] proposed an approach that facilitates continuous mission plan adaptation. In principle, the approach maintains a window for actions that have the potential to be performed in response to the real-time circumstances instead of following commonly utilized mission planning techniques that are based on providing a complete plan of the mission during the initial phase. The proposed approach continuously reassesses the status of the environmental conditions and identifies the changes in the platform capabilities and from the list of probable actions select those that maximize the outcome of a utility function.

Karimoddini et al. [48] presented a hybrid supervisory control framework that facilitated 3D leader-follower formation control for UAVs. The framework uses a spherical abstraction of the state space in addition to multi-affine functions. The hybrid system uses a finite state model that is bi-similar to the original continuous-variable dynamical system in addition to a logic supervisor in order to provide necessary control over the formation of the UAVs during the mission.

Guerriero et al. [49] utilized a distributed version of self-coordinated and cooperative UAVs in scenarios based on searching for a Points of Interest (PoI) in a time restricted fashion. The authors considered two strategies of a rolling horizon and the ε-constraint for optimizing the objectives of the study (maximizing customer satisfaction, minimizing UAVs travel distance, and minimizing the number of utilized UAVs in the mission). The ε-constraint method assumes complete knowledge of the environment and the PoI, and their related timing factor (e.g., born, starting time, and end time for each PoI), which appear in the environment during the mission, and address the multi-objective optimization problem by providing a set of ε-constraints for each objective separately at each state/iteration. The rolling horizon strategy is designed based on the availability of partial information about the location and the related time instant of each event, and at each iteration, the UAV routing task is performed based on the partial information available at that iteration.

Saska et al. [50] investigated leader-follower-based formation control problem and proposed an approach based on utilizing pairs of virtual leaders that are controlled by an optimization process. The process uses a model predictive control approach, and the resulting solution of the optimization process includes a complete control plan in addition to controlling inputs of individual UVs. The use of the virtual leaders as suggested by the authors increased the maneuverability of the UVs while helping to stabilize the shape of the formation.

Keller et al. [51] utilized B-spline curves to provide onboard time-optimal trajectory management. The B-Spline is developed using a set of algorithms that transcribes sets of points to a continuous path. The approach post-processes the point-by-point path planner's output to generate minimal representation aiming to provide minimum execution time, dynamic adaptation to acquired sensory-feedback, and adherence to mission constraints. The approach extends B-Spline to include features such as real-time trajectory interruption and redirection.

Kopeikin et al. [52] studied the networking aspects of mission management and highlighted the importance of being able to maintain the communication relay between a fleet of UAVs and the ground control station. The networking problem is handled through distributed task allocation that enables UAVs to play the role of network relayers to maintain communication to the base. This is facilitated by coupling task assignment and relay creation processes, resulting in solutions that are capable of addressing realistic network communication dynamics (path loss, routing, and stochastic fading). The networking aspect and its constraints on mission planning and management is also investigated by Manousakis et al. [53], who note that defining the required networking performance is part of the mission planning process. The study focused on utilizing a network management system called TITAN (Tactical Information Technologies for Assured Networks) in order to manage the networking requirements of the mission, which generates maintenance policies based on the mission goals and the network plan. The proposed mechanism uses the policy-defined actions when a network contingency appears to restore the network performance and utilize the TITAN system if the pre-defined policy failed to restore the network performance. In this context, the TITAN role is to dynamically re-plan the network to fulfill the mission objectives. The dynamically re-planning process is facilitated using a Mission to Policy Translation (MPT) mechanism that utilizes the existing resources in each stage in a way to maintain the satisfaction of the mission objectives.

Doherty et al. [54] proposed a conceptual architecture that is based on the unifying concept of delegation. The architecture incorporates four components of: a mission specification language based on Temporal Action Logic (TAL); a distributed structure based on Task Specification Trees (TSTs) that is utilized for defining single and multi-platform missions; an automated planner that is capable of supporting mission specifications (single or multi-platforms); and a multi-agent architecture based on a delegation concept that supports collaborative missions and is formalized as speech act. The authors discussed some particular benefits of their proposed architecture such as being generic to be used on different robotic platforms as a result of separating the architectural extensions and the robotic system's legacy, and the ability to instantiate mission specification and planning approaches to different application areas which is useful in heterogeneous robotic systems.

Adams et al. [55] investigated the potential of Cognitive Task Analysis (CTA) and Cognitive Work Analysis (CWA) procedures in the development of autonomy algorithms for UAVs involved in Wilderness Search And Rescue (WiSAR) missions. The UAV-based WiSAR are complicated missions in which a hierarchy of human decision making exists in the following steps: first, the UAV searches the unknown environment for signs of a missing person and reports the findings in real-time to the UAV operators; secondly, the UAV operators pass the live video feeds to the video analyst person to investigate signs of the missing person; ultimately, the video analyst informs the incident commander about the new evidence and directs ground searcher personals toward areas with the highest probability of success. The authors argued that the introduction of UAV into WiSAR mission does not convert the manual human-based search process to an autonomous search, but it changes the manual process to a cognitive process with the advantage of minimizing the physical search process in addition to providing cognitive processes to the incident commander in order to better manage the search. The study suggested combinations of goal-directed task analysis and CWA for improving the UAV design to better support the WiSAR mission and utilization of CTA for enhancing the UAV's autonomy.

Mufalli, Batta, and Nagi [56] tackled the problem of sensor assignment and routing in a UAV-based study. The study is emphasized on a mathematical approach that facilitates simultaneous sensor allocation and routing using CPLEX (for small problems) and some heuristics (for larger problems). The authors considered constraints such as UAVs with interchangeable sensors and payload weight in addition to travel time variability of the UAVs, limitation in fuel consumption, and time-dependent target visiting tasks.

Binetti, Naso, and Turchiano [57] considered the problem of allocating tasks with heterogeneous natures in which some are considered as critical for the overall success of the mission. The authors employed a Decentralized Critical Task Allocation Algorithm (DCTAA) with a three-phase iterative structure that in the first phase allocates tasks either by adding or forcing, in the second phase provides procedures for conflict resolution and in the third phase remove tasks from agents that overstepped their task capacity.

Andrews, Poole, and Chen [58] employed phase mission analysis for the quantification of system reliability in UAV's mission planning as part of the decision-making process. The investigation is focused on generating a plan for a mission that has the highest probability of success through minimizing the probability of failure in each phase of the mission. The role of the mission analysis phase is to identify the likelihood of failure of the mission in each phase subject to completing all previous phases successfully. The proposed approach is tailored based on carrying out a majority of the processing in advance in off-line mode through the use of a library of probable failure causes of all potential phases.

Petillot et al. [59] presented an ontology-based framework to move from a way-point representation of mission planning towards goal-based planning, while including multiple collaborating vehicles scenarios. The authors introduced concept of adaptive autonomy as a more suitable approach rather than re-planning the mission for dealing with the changes in the environmental conditions and vehicles capabilities. The adaptive autonomy can be achieved through the use of SA procedures that provide a higher level of understanding of the environmental dynamism and detection of changes and events that contradict with the initial situations that are considered when the plan is generated.

## 6    The AUV-based Mission Planning and Mission Management Systems

Growing attention has been devoted in recent years on the autonomous application of AUVs in performing different tasks in dynamic and continuously changing environment increasing the ranges of missions and promoting vehicles autonomy to handle more extended missions without supervision [6]. Williams [60] developed an adaptive track-spacing algorithm for an AUV optimal route planning approach used in mine countermeasures mission in which the operation is framed in terms of maximizing underwater mines detection. No onboard processing capability is incorporated in the route planning resulting in non-adaptiveness of the approach to dynamic changes.

Niaraki and Kim [61] modeled vehicle route planning with a multi-criteria decision-making mechanism and applied a generic ontology-based architecture taking advantage of a hierarchical analytic process to determine the choice of criteria for using an impedance function in the route-finding algorithm. In the proposed strategy, the domain-specific ontology provides a foundation for maintaining or extending the domain knowledge. An autonomous goal-based planning approach is presented by Patron and Ruiz [62] using a knowledge-based framework, and service-oriented architectures. The authors have shown that this methodology enables domain-independent planning and capability of discovery for complex missions without experts' involvement in which the system can handle poor and intermittent communications by predicting the intent of the other platforms to accomplish goal selection. The problem associated with knowledge-based (ontology-based) systems is that consistent information about the correct action and environmental situations is not always available.

Similar to the UAV-based mission management systems, many researchers approach the mission planning and task assigning problem by transforming the search process to waypoint tracking graph search problems. In such problem, tasks can have different meanings including finding the optimal route between two points, performing certain functions such as collecting samples, performing certain maneuvers, mapping and so on. These tasks can be assigned to waypoints or can be distributed across the operation area of the mission and within the route from one waypoint to another. In these approaches, the real-time performance of the application are usually overshadowed by the growth of the graph complexity or problem search space. Growth of the search space increases the computational burden that is often a problematic issue with deterministic methods such as Mixed Integer Linear Programming (MILP) proposed by Yilmaz et al. [63] for governing multiple AUVs. The meta-heuristic algorithms, however, offer near-optimal solutions in faster time and with better scalability comparing to a deterministic method such as MILP when multiple tasks are to be performed. For example, in meta-heuristic approaches, the definition of tasks can be converted to objectives for optimization which provides the opportunity to address them within the context of a) multi-objective optimization problem, b) single but serialized objective optimization problem, or c) multi or single objective optimization problem for which predefined and hard-coded heuristics are to be used to assure the delivery of a portion or majority of the objectives resulting in a computationally less expensive problem solving mechanism. Increasing the number of tasks in such system is equivalent to a) increasing the optimizing objectives, b) small alteration to objective serialization mechanism, and c) inclusion of new heuristics for delivering the newly defined tasks. This flexibility in handling inclusion of additional tasks attest to scalability of these methods. Alternatively, in this context, increasing the number of tasks can, in some cases, refer to increasing the operational envelope of the mission by adding new waypoints and passages in between them. While such circumstances impose a degree of computational burden on meta-heuristic methods, the computational costs on MILP method is known to be more severe since in this method the computational time increases exponentially with the problem size.

Several studies also investigated the advantages of meta-heuristics methods on mission management in terms of vehicle routing and task assignment. A comparative approach employing Dijkstra and Genetic Algorithm (GA) is used by Sharma et al. [64] to address the vehicle routing problem in a graph like operation network. The evaluation results affirmed that both algorithms provided the same solution but with different execution time as Dijkstra's is time-consuming compared to GA. A large-scale route planning and task assignment joint problem related to the AUV activity has been investigated by MahmoudZadeh et al. [65] by transforming the problem space into an NP-hard graph search context. The approach used heuristic search nature of Particle Swarm Optimization (PSO) and GA to find the best waypoints order to address AUV's task assignment and risk management in a large-scale static terrain. This approach is extended in [66] to a semi-dynamic network using Biogeography-Based Optimization (BBO) and PSO algorithms, where the location of the waypoints and the topology of the network is considered to be dynamic and uncertain. The merit of the proposed solution by [65, 66] is being independent of the graph size and complexity.

Woodrow et al. [67] reported the development of concepts that are designed to improve the level of AUVs' autonomous planning and mission management. The study is focused on development of *i)* an intuitive user interface that moves away from specifying missions as a detailed scriptive approach toward a set of military goals, *ii)* an autonomous onboard mission planning/re-planning software that help AUVs to compensate for the changes in the environment, mission goals, or status that demand mission re-planning, and *iii)* transit planning software that makes the autonomous route planning possible with respect to the environmental conditions, possible risks, and user defined weights. The planning module of the package employs a hierarchical approach in which first a plan based on simple task models is generated, and the simple plan is refined in later levels to incorporate more detailed task plans. The re-planning level is split into two layers of mission re-planning and task planning layers. The developed transit planning software considers some environmental and risk-related factors for planning a transit path. These factors include *i)* Time-varying and non-uniform subsurface currents, *ii)* Time-varying water levels, *iii)* Areas or times of high physical risk, *iv)* Exclusion zones, and *v)* Risk of detection.

Bian et al. [68] studied the implication of AUV for ocean surveying tasks in dynamic and partially known environments. The authors suggested an on-board hierarchical architecture that utilizes the agent-based distributed autonomous control structure. Mission planning/management are addressed through the use of discrete events driven Petri Nets (PN) formalism. The use of this hierarchical PN based mission management and control system allows the AUV to process multiple discrete events. In this method, the highest level PN models the execution of the mission phases while the lower level sub-PNs provide mission control. The architecture is evaluated with lake trials in which promising results are achieved.

Albiez, Joyeux, and Hildebrandt [69] proposed an adaptive AUV mission management architecture that utilizes a plan management mechanism that uses both predictive and behavior-based approaches to control the AUV and handle under-informed situations. The introduced hybrid reactive/deliberative architecture uses behavior-based methods in order to manage the tasks and uses an elaborate plan manager to control the deployment, activation, and deactivation of the behaviors in order to maintain the progress and fulfill the mission. The architecture includes blocks of sensor processing, vehicle management and safety, behavior pool, and plan management. The feasibility of the architecture is assessed using AUV AVALON in a pipeline leak detection scenario.

The implication of intelligent decision-making algorithm for AUV that is tasked to perform in-depth oceanic survey autonomously is studied by Bian et al. [70]. The study considered environmental constraints such as forbidden zones, unsafe zones, and obstacle areas. An intelligent decision algorithm constitutes a global path optimizer, and a speed optimization method in combination with a hierarchical on-board architecture is utilized for executing the mission. The decision algorithm takes into account the environmental constraints in addition to mission information and AUV's status/restrictions in order to generate the mission commands. The mission management architecture utilized in the study is based on increasingly detailed PN in which the separation of mission operations in different PNs increases the feasibility of adding new operations. This study is advantageous due to *i)* synchronous event handling, *ii)* programmability, *iii)* reactivity, and *iv)* adaptability for the utilized mission management architecture. In a similar scenario, Rajan et al. [71] dealt with abstracted diagnosis and failure recovery aiming to provide solutions for the robust, sustained presence of AUV in harsh underwater missions such as in-depth oceanic surveys. Investigating and analyzing current AUV operations model, the study identified some critical issues to be addressed in the design of AUV-based autonomous mission management system including: failure to turn up on time; uncertainty in onboard computation; sensor failure conditions; impaired mobility; lack of built-in SA; limited adaptability to opportunistic events.

Pfuetzenreuter [72] tackled the AUV's mission re-planning complexity in surveying in-depth oceanic scenarios in which communication with human operators and requesting mission update was impossible. Pfuetzenreuter described the proposed architecture in functional modules of mission control, mission plan handling, mission monitoring, mission re-planning, and chart server. In this approach, the mission monitoring module is responsible for mission/plan initiation and observing mission execution while mission re-planning module is responsible for detecting the necessity of plan modification and generating mission updates that optimize the outcome concerning the mission objectives. The chart server module is tasked to check the modified plan against digital charts to confirm its consistency with terrain constraints, and in case of detecting violation(s) the system considers alternative plans.

Another recent attempt for mission planning in the scope of underwater operations is done by MahmoudZadeh et al. [73], in which a reactive control architecture is introduced to provide a higher level of decision autonomy for an individual autonomous vehicle to accomplish multiple tasks in a single mission in the face of periodic disturbances, turbulence, and uncertainties. The system incorporates two execution layers, deliberative and reactive, to satisfy autonomy requirements in both high-level mission scheduling and low-level motion planning for generating prompt actions/reactions. The authors implemented an efficient synchronization scheme as an integral part of the architecture to allow coordination of modules in different layers of the system.

MahmoudZadeh et al. [74-76] introduced a new perspective to the AUV's mission planning problem by developing a synchronous scheme for decision making of AUV missions. These studies provided vehicle's decision autonomy through designing a higher-level decision-making system for ordering the execution of the tasks and guiding the AUV toward the targeted location; and using a lower-level motion planner as an action generator to ensure a safe and efficient deployment according to vehicles awareness of the situation. They mostly focused on managing mission time (vehicle battery lifetime) and increasing mission productivity. The performance and accuracy of two higher and lower level modules were tested and validated using some population-based meta-heuristic approaches. Although the research emphasized a new point of view to autonomous mission planning, some details such as effect of water current force (as an effective factor on underwater operations) on AUV's deployment is not addressed.

Later on, MahmoudZadeh et al. [77, 78] provided a mature and completed version of previous studies [74-76] a modular control architecture was developed consisting of two deliberative and reactive execution layers. They contributed an autonomous interactive modular approach where the deliberative layer manages the concurrent execution of several tasks with different priorities and the reactive layer manages real-time reactions and performs a quick response to critical events. They emphasized a more realistic underwater environment comprising dynamic ocean current and various uncertain moving/afloat objects, while the motion planner also is integrated with dynamic re-planning capability to deal with ocean dynamicity. In this approach, the deliberative and reactive layers operate concurrently with interactive back feeding of changes in the environmental conditions, goals, plans, and situations. These automatic functions are executed in the background and enhance AUVs self-management characteristics. The subject area is one that is of importance, as the authors point out, steps to reduce the reliance on expert operators that contribute to the scalability of AUV operations and also improve reliability and repeatability of operations. The architecture offers a degree of flexibility for employing divers sorts of algorithms and can perfectly synchronize them to have a stable performance. However, this research is developed and evaluated through some simulations, which may not be sufficient for real-world experiences.

Brito et al. [79] presented the key obstacles for the integration of adaptive mission planning techniques for AUVs based on the exchanged opinions between experts in the field and the results gathered from some questionnaires. The study identified various reasons for the failure to adopt adaptive mission planning software including:

1) Technology is not understood (20.7%)
   - Technology is not well explained (55.4%)
   - Technology is too complex (44.6%)
2) Uncertainty with regards to vehicle responses (39.7%)
   - Insufficient demonstration trials (65.8%)
   - Lack of risk assessment (34.2%)
3) Technology is too expensive (21.5%)
   - Development costs are not tangible (50.7%)
   - No predefined development life-cycle (49.3%)
4) Benefits are not significant (15.0%)
5) Uncertainty with regards to legal limitations (3.1%)

Based on the gathered opinion from experts in the field, in the first level, uncertainty with regards to vehicle response is recognized as the most probable cause of failure to use adaptive mission planning software in AUV community. Within the second level, insufficient demonstration trials (26.1%) and the lack of risk assessment (13.5%) are considered as the probable cause of the failure to use these type of methodologies. The following section discusses the complications arises by having a swarm of UVs participating collaboratively and cooperatively in various robotic scenarios.

## 7    Autonomous Mission Planning and Management Systems for Swarm Robotic Scenarios

Developing swarm based solutions for teams of robots that tackle problems through cooperation, collaboration, and coordination have several advantages over the use of singular robots with high abilities, skills, and intelligence that are very expensive and complicated. There are several similarities between UAVs, and AUVs in the context of mission planning and autonomous decision making, so we are also going to discuss the way that these problems are addressed over the multiple vehicles' collaborative operation.

Dominik et al. [80] implemented a route-plan-decision-maker for multi-agent transportation planning, where routing problem is considered with specific time windows, cluster-dependent tour starts, and the agent decides between distributing orders to customers, traversing edges, competing vendors, increasing production, etc. The outlined study focused on dividing jobs between multiple robots while they are moving towards a destination point and managing the loose (break down) of any of robots in the group by reassignment of its tasks into the closest robots. However, in this study, the environment is assumed to be ideal, while there are many other uncertainties and considerations in real world. Certainly, many problems will be revealed when it comes to practical applications especially in the case of autonomous unmanned robots.

Geng et al. [81] utilized a combination of GA and Ant Colony System (ACS) for mission planning of a team of autonomous UAVs that get engaged in urban surveillance missions. In the study, GA is utilized during the first stage (mission planning) to identify the minimum number of required UAVs to be engaged in the mission in order to provide constant surveillance over the 3D environment. ACS is utilized in the second stage (path planning) to minimize the required changes in the traveling altitude of each UAV in the team while following a path close to the shortest possible distance.

The employment of a swarm solution, however, requires a change in the way that a mission is planned and furthermore requires a proper task allocation between the multiple UAVs. In such situations, the teammates of any UAV can represent potential obstacles and any change in the pre-planned trajectory/ attitude of a UAV in the team, which provides a higher level of cautiousness in order to prevent the collision with other friendly UAVs. Planning problems that are concerned with coordinating actions of multiple assets subject to failure are also referred to as "Multi-Agent Health Problem". Raghvendra et al. [82] counted redundancy, faster completion time, and ability to cover larger areas as some of the advantages of using multiple UAVs in complex missions over the use of single UAV. They addressed this problem through the use of a temporary command control that allows any UAV in the team to take over the control of the team and ask other UAVs to readjust their position in order to allow sufficient maneuvering space/distance whenever a re-planning is essential. The approach keeps the human operators in the loop by allowing them to overrule any action if it is necessary.

Jameson et al. [83] reported the development of a general architecture that allows a high degree of autonomy within UAVs collaborative team operations. The developed architecture takes advantage of an onboard mission planning module, which considers aspects such as intelligence, collaboration, awareness, responsiveness, and agility. The module works in a hierarchical manner that plans missions at the highest level (for multiple teams of UAVs) and in mid-level generates plans for each team in the swarm of UAVs and individual UAV at the lowest level. The generated plans consider factors such as: *i)* mission objectives and constraints, *ii)* resource allocation, *iii)* payload configuration for various objectives, *iv)* collaborative use of gathered information, *v)* communication, *vi)* routes that satisfy the plan, and *vii)* target planning. The collaboration components consider factors such as sharing information, role and responsibility allocation, task coordination, dynamic team forming, and interacting with external assets. The proposed architecture in [83] takes advantage of Lockheed Martin's MENSA technology for contingency management. The MENSA incorporates routines that make it capable of identifying plan dependencies and constraints, and it explores conditions that violate these constraints.

Hoshino and Seki [84] studied interaction force and behavior regulation rule method on the coordination problem of multiple robots in congested systems and proposed amplification of the robots' velocities through the modified behavior regulation. This is achieved by using interaction forces between robots and reducing robots' velocities in certain areas. In addition, the study proposed generating direct control of the robot's velocity in the behavioral dynamics. The methodology is evaluated in simulations and is proved to be effective for the coordination of multiple robots in a congested systems with bottlenecks. Kudelski et al. [85] introduced a simulated environment called RoboNetSim capable of modeling network constraints for multi-robot scenarios. The simulator provides realistic networking efficiency measurements for generated scenarios compatible with the real-world environments.

Yamamoto and Okada [86] introduced a swarm controlling approach inspired by the appearance of congestion in Crossing Pedestrian Flows (CPFs) in high-density urban areas. Pedestrians avoid collision with other pedestrians in CPF scenarios using self-organized phenomena. The continuum model of CPF is proposed by Yamamoto and Okada with the ability to generating density factor from dynamic changes of the congestion. The authors improved the average velocity by utilizing a control method. The proposed approach considers the diagonal stripe patterns shaped by the crossing pedestrians and takes advantage of the dynamic interactions among these diagonal stripe patterns in order to guide the moving person/object in the flow. The feasibility of the approach is assessed through a simulation study.

De Lope et al. [87] investigated heterogeneous task selection in multi-robot coordination. The principle idea developed in the study is based on generating decentralized solutions by allowing robots individually and autonomously selecting tasks with the constraint of maintaining optimal task distribution. The constraint is facilitated using learning automata-based probabilistic algorithms and the Response Threshold Model (RTM). The potential of this approach is assessed through simulation study using random and maximum principle approaches for task selection. The results indicated that the maximum principle is a better fit for the task selection process. Furthermore, it is concluded that the RTM is robust to noise and it is more time-intensive in comparison with the learning automata-based probabilistic approach. The results indicated the suitability of this method for addressing issues relating to task allocation within a swarm of robots in the absence of any central task scheduler unit.

Combination of high-level structures and a bottom-up approach is proposed by Zhu et al. [88] for navigating a swarm of robots. The use of a bottom-up (low-level) layer provides some degree of autonomy and intelligence for individual robots, in addition to ensuring their independence, while the convergence in complex and critical circumstances is guaranteed via the use of a higher level layer of intelligence. The advantage of the proposed bottom-up approach in comparison with the conventional top-down hybrid architectures is demonstrated by the ability to incorporate various algorithms designed for handling certain circumstances. In contrast to the discussed multi-layer level of intelligence in [88], Pimenta et al. [89] studied an approach that forbids individuality and independent decision making within the swarm and utilizes a single software/algorithm to control the swarm. The principle idea in this study is to direct the swarm of robots (as a whole) towards the region of interest. This is advantageous since the swarm is no longer dependent on individual members and the generated solution is robust concerning dynamic alternation of the swarm population (addition and deletion of swarm members). The Smoothed-Particle Hydrodynamics (SPH) simulation technique is utilized to generate an abstract representation of the swarm, which provides a loose way for controlling the swarm. The approach is evaluated through simulation and experimental studies; the results indicate the efficiency of the approach in dealing with a high number of agents (more than 1000) in simulation and also smaller swarm sizes.

Fukushima et al. [90] studied the formation of a swarm of UVs and proposed a formation control law with the ability to convert the optimal control problem to a mixed integer quadratic programming problem. This conversion is facilitated by utilizing feedback linearization with a Model Predictive Control (MPC). Besides, a Branch-and-Bound (B&B) method is applied to handle the obstacle avoidance. The results achieved from numerical experiments showed significant computation time reduction.

Hemakumara and Sukkarieh [91] discussed a system identification method based on the dependent Gaussian process. The system is designed to handle modeling constraints of stability and control derivations of UVs. Dependency on prior knowledge of the model structure and not being capable of addressing a wide flight envelope are two of the considered modeling constraints. In the proposed method, the control derivatives and system stability are addressed by capturing cross-coupling of input parameters. The approach is evaluated through simulation experiments using a highly coupled oblique wing vehicle and delta-wing UAV platforms. The results illustrated several advantages over traditional methods including having efficient identification of coupling flight modes, a higher level of robustness to uncertain estimations, and being applicable to wider flight envelopes.

Atyabi et al. [92, 93] studied the problem of controlling a small swarm of simple robots in search and rescue scenarios with various degrees of complexity (dynamism, time dependency of rescue tasks, heterogeneous skill distribution among swarm members). The study is focused on investigating the potential of two PSO-based approaches of Area Extended PSO (AEPSO) and Cooperative AEPSO (CAEPSO). AEPSO and CAEPSO shared some commonalities in terms of their heuristics with CAEPSO having the extra ability to handle a combinatorial type of noise and learn from previous learning sessions. AEPSO performed a better local search in comparison with a variety of approaches including basic PSO, random search and linear search in both time-dependent and dynamic scenarios. The robustness of the AEPSO and CAEPSO to noise and task-time-dependency is considered to be the result of having better cooperation between agents achieved by knowledge sharing and balancing the use of essential behaviors of the swarm (exploration and exploitation).

Acknowledging the relevant studies in the scope of ground and aerial vehicles, the mission management becomes even further challenging considering the severity of the underwater environment. The underwater environment is generally a vast 3D area, which is rarely populated with obstacles and usually unknown in advance [30]. The challenges associated with such an uncertain environment becomes even more significant in long-range missions with enlargement of the operation area. Restrictions of apriori knowledge about later conditions of the environment attenuate AUVs' autonomy and robustness.

In the scope of underwater operations, Yan et al. [94] developed an integrated mission planning strategy to serve the AUVs' routing and tasking problem, while minimizing total energy cost in the presence of ocean currents was the main objective of the study. Iori and Ledesma [95] proposed a behavior-based controller coupled with waypoint tracking scheme for a double vehicle routing-task planning problem with multiple stacks, where the routing problem is modeled with a Double Traveling Salesman Problem with Multiple Stacks (DTSPMS) for a single-vehicle pickup-and-delivery problem. They studied the performance of three branch-and-cut, branch-and-price, and branch-and-price-and-cut algorithms on a wide family of benchmark test samples with different graph complexities. Chow [96] proposed an approximated algorithm based on a k-means method to address task allocation problem that minimizes the travel time for multiple AUVs considering the presence of constant ocean current. The approach combined the AUV dynamic model with an adjusted Dubins model including ocean currents for AUVs motion along the route with a constant velocity. However, the problem is considered in a 2-D environment that would be problematic when the method is applied to accomplish the similar task in a 3-D environment. Moreover, they considered a static ocean current rather than the variable current, which is not prevalent in a real-world application.

Zhu et al. [97] integrated a Self-Organizing Map Neural Network (SOM-NN) with a velocity synthesis approach for solving multiple AUVs' dynamic task assignment and path planning problem. They considered multiple target locations in a 3-D environment in the presence of dynamic ocean current. Although the approach in [97] is capable of keeping AUV on the desired track during the mission, the environment is assumed ideal, while there are many other uncertainties and considerations in the real world, such as static or moving obstacles and so on.

Atyabi et al. [98, 99] reported the development of a system called Strategy Over-Lord (SOL) designed to oversee the moment-to-moment operations of multiple teams of heterogeneous UVs with the ability to detect near future events and proposing required changes to the robot's behaviors to the operators. Besides, based on the observed changes in the environment the SOL system is capable of offering five different solutions based on factors or any combinations thereof such as distance, associated direct and indirect risk of an enemy encounter, maneuverability of the proposed path, and network coverage. The system is capable of prioritizing between the proposed solutions and also designed in a way to allow the operators to interfere with the made decisions by SOL by rejecting the proposed actions. The authors argued that it is expected that the system would reduce the workload of the operators and improve the overall achievements of robots due to capabilities such as: *i)* announcing vital information that might be missed by operators as a result of the large quantity of information presented by robots and on-going changes in the environment; *ii)* reducing the workload of operators by providing possible solutions/actions for the announced problem (subject to operator trust in the SOL's solutions and near future prediction capabilities); and *iii)* dynamically readjusting the robots or the teams of robots strategy based on the current or near the future status of the environment (assuming that operators allow the actions to be passed to the robots).

Summary of the discussed approaches to UAV, AUV, and swarm-based mission planning and mission management systems is provided by Tables 1 to 3 to conclude the outstanding findings of the given literatures.

## 8 Conclusion

This article provided a review on some of the most recent developments in the field of mission planning and mission management with a focus on Arial and Underwater vehicles. The article discussed the challenges faced by autonomous AUV and UAV platforms and provided an in-depth discussion on existing measures of autonomy by providing a case study scenario for comparison of these autonomy measurement approaches with human controlled, semi-automated and fully automated UVs highlighting in ability of these methods for adequately distinguish various levels of autonomy and the confusions their use can cause which is also reflected in 2012 final report of the task force to DoD [23].

Following the discussions on how to measure level of autonomy in UVs and the precision of such measurements, the article presented existing definitions for mission planning and management systems and further described components such as situation awareness and cognition while presenting various innovations introduced by the robotics and autonomous systems society. The state of the art designs and applications of autonomous mission planning and management systems used in UAV and AUV applications are discussed and their pros and cons are counted. The study is concluded with applications of mission planning and management systems in multi-UV domain (swarm robotics) and the existing innovations in the field. This is an important direction since the full incorporation of autonomy and onboard intelligence in a level that can guarantee timely and correctly achievement of mission objectives provides the opportunity and future demand on scenarios where

1. More complex and demanding missions in which more than one UVs are required for delivering mission objectives.

2. Multiple missions are expected to be conducted simultaneously with the same mission planning and management system and due to lack of limitation in resources in terms of access to UV equipment, the existing resources are to be shared across multiple concurrently running missions.

3. Multiple fleets of UVs are to be deployed to multiple missions simultaneously.

Given the complexity of these scenarios and the possible need to access/use sensitive and costly sensors, it is reasonable to believe that a cost efficient autonomous mission planning and management system will incorporate routines and heuristics allowing the use of dynamic swarms where skills and capability (sensors, UVs, networking equipment, and so on) are to be shared across multiple swarms that each is involved in a portion of a) a large mission or b) separate missions that are within vicinity of each other.

Although several studies are conducted for developing methodologies that fulfill the requirements of the mission planning and management problems, the role and impact of the human operators are mostly ignored. That is, it is common to blame the mission planner and mission manager for the possible failure of the mission while operators also have a high impact on the moment-to-moment progress of the mission. This issue becomes more critical in scenarios where several unmanned vehicles are being controlled by very few operators (e.g., two operators) in environments for which limited information exists. In such scenarios, the dynamism of the environment requires constant changes to be applied to the pre-planned mission which also requires having a higher level of SA by UVs and operators in addition to having a higher level of workload on operators which ultimately results in their failure in terms of making proper adjustments to the plan.

Another aspect of involvement of human operators in delivery of missions is the entanglement of mission progress and the type and frequency of information being reported to operators. This issue becomes more apparent in multi-swarm multi-mission scenarios discussed earlier. High level of mental workload caused by frustrating task of following the operational flow of multiple UVs, fleets and missions can be worsened by using inappropriate and excessive flow of on screen information. This issue links the future development of complex multi-mission multi-swarm mission planning and management systems with Human Computer Interaction (HCI) field where such facets are studied and practiced in software designs and architectures.

While one solution to complex problem of multi-swarm multi-mission platforms is to consider cooperation of multiple operators, several considerations are to be in place to guarantee successful collaborations such as how the communication between the two or more operators to be handled? what are the rules for collaborations between operators? Do each operator take responsibility over one or more UV fleet(s), mission(s), or task(s) or different mechanisms are to be used for guaranteeing constructive collaboration? what are the responsibilities of each operator? how operators are to be matched with each other? what degree of onboard automation and intelligence to be used to maintain the trust of operators on UVs and fleets and to minimize their operation interruption by operators? and several other factors that are beyond the scope of the current report.

Considering the aforementioned scenarios and problems, it seems necessary to develop adaptive mission planner and management systems that not only automate the plan readjustments but also reduces the workload (both mental and physical) on operators while increasing their SA.

## Biographies:

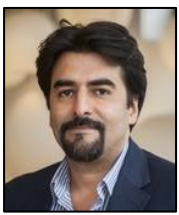

**Adham Atyabi** received his PhD from Flinders University of South Australia in 2013. He is currently Technology Lead in Seattle Children's Innovation & Technology Lab and Senior Postdoctoral Fellow at University of Washington. His research interests include Autism Spectrum Disorder, Cognitive Neuroscience, Computational Neuroscience, Computational Intelligence, Eye Tracking, Image & Signal Processing, and Swarm and Cognitive Robotics.

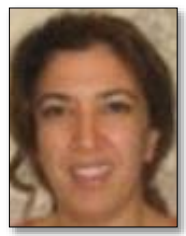

**Prof. Nefti-Meziani** holds Doctorat D'etat in robotics and artificial intelligence and is Director of the Centre for Autonomous Systems & Advanced Robotics, and Chair of Robotics at the University of Salford. In this role, she leads a multidisciplinary team of 6 academics and 12 researchers in automation, robot/machine design, dexterous end effectors, legged robots, soft robotics, biologically-inspired robots, haptics/telepresence, physical human-robot interaction, rehabilitation robotics, cognitive robotics, uninhabited autonomous systems.

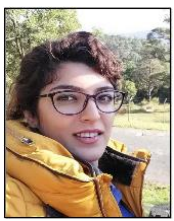

**Somaiyeh MahmoudZadeh** received her PhD from Flinders University of South Australia in 2017 in Computer Science (Robotics and Autonomous Systems). She is currently acting as Postdoctoral Research Fellow in Faculty of IT, Monash University. Her area of research includes computational intelligence, autonomy and decision making, mission planning, situational awareness, and motion planning of autonomous underwater vehicles.

**Table.1.** Summary of recent approaches to UAV-based Mission Planning and Mission Management Systems

| Techniques | Ref | Highlights |
|---|---|---|
| Autonomous Control executive (ACE) as part of TRAC mission management architecture | [42] | This method is advantaged to issue mission segment commands, monitor segment completion, react to unexpected events, manage start-up and shutdown. |
| Intelligent Mission Management (IMM) accompanied with: CDE and Intelligent Agent Architecture | [43] | This method allows UAVs to redirect the flight according to the changes in environmental conditions. The CDE (Collaborative Decision Environment) system provides SA, collaboration, and decision-making tools for UAV. The system uses an automated planning tool of MAPGEN as a ground-based decision support system that allows human operators to make necessary changes. |
| Holistic Contingency Management | [44] | The proposed method provided multi-level assessment; plan-based assessment; capability-based assessment; predictive assessment; and team-based assessment. The authors investigated the importance of contingency management in mission planning to compensate and recover from problems resulting from unanticipated events. |
| Cub-Space (C-Space) Mission Planning | [45] | The approach contained modules to satisfy SA, autonomous mission planning, and collision avoidance. Mission planning was handled through employing the C-Space method to represents the world in 3D cubes that are marked either free or occupied, where the set of free cubes drive the UAV toward the goal by taking the shortest distance. |
| Mission Displays for Autonomous Systems (MiDAS) with a Cognitive Work Analysis (CWA) | [46] | The study used MiDAS to direct the attention of the operators toward the aspects of the plan that are likely to influence the automation system's functioning and the applied CWA to analysis of the Intelligence, Surveillance, and Reconnaissance of the system. |
| Real-Time Adaptive Mission Planning Approach | [47] | The approach facilitates continuous mission plan adaptation and maintains a window for actions that have the potential to be performed in response to the real-time circumstances. The method continuously reassesses the status of the environment, identifies the changes in the platform capabilities, and selects an action from the list to maximize outcome of the utility function. |
| Hybrid Supervisory Control Framework | [48] | The framework facilitated 3D leader-follower formation control for UAVs applying a spherical abstraction of the state space in addition to multi-affine functions. The system uses a finite state model that is bi-similar to the original continuous-variable dynamical system accompanied with a logic supervisor to provide necessary control over the formation of the UAVs during the mission. |
| Rolling Horizon and ε-Constraint | [49] | Considered two strategies of a rolling horizon and the ε-constraint for minimizing UAVs travel distance and minimizing the number of utilized UAVs in the mission. The ε-constraint method assumes complete knowledge of the environment and the points of interest. |
| Optimization-based Virtual Leaders | [50] | The process uses a model predictive control approach, and the resulting solution of the optimization process includes a complete control plan in addition to controlling inputs of individual UVs. The use of the virtual leaders increased the UVs' maneuverability while helping to stabilize the formation shape. |
| B-Spline point-by-point Path Planner | [51] | The approach post-processes the point-by-point path planner's output aiming to provide minimum execution time, dynamic adaptation to acquired sensory-feedback, and adherence to mission constraints. |
| Coupling Task Assignment and Relay Creation Processes | [52] | Investigated networking aspects of mission management and highlighted the importance of being able to maintain the communication relay between a fleet of UAVs and the ground control station. The networking problem is handled through distributed task allocation that utilizes UAVs to play the role of network relayers to maintain communication to the base. |
| TITAN (Tactical Information Technologies for Assured Networks) | [53] | The study manage the mission networking requirements, which generates maintenance policies based on the mission goals and the network plan. The dynamically re-planning process is facilitated using a Mission to Policy Translation (MPT) mechanism. |
| A Conceptual Mission Planning Architecture | [54] | The architecture incorporates four components of: a mission specification language based on temporal action logic; a distributed structure based on task specification trees to define missions; an automated planner to support mission specifications; and a multi-agent architecture based on a delegation concept that supports collaborative missions and is formalized as speech act. The proposed architecture is generic and applicable on different robotic platforms as a result of separating the architectural extensions and the robotic system's legacy. |
| Goal-directed Cognitive Task Analysis (CTA) and Cognitive Work Analysis (CWA) | [55] | The study investigated the potential of CTA and CWA procedures in the development of autonomy algorithms for UAVs involved in WiSAR missions, where the conventional manual search process is handled by an autonomous cognitive process. The study suggested combinations of goal-directed task analysis and CWA for improving the UAV design to better support the WiSAR mission and utilization of CTA for enhancing the UAV's autonomy. |
| CPLEX and some heuristic methods | [56] | The study tackled the UAV problem of sensor assignment and routing through a mathematical approach to facilitate simultaneous sensor allocation and routing using CPLEX for small problems and heuristics for larger problems. The authors considered constraints such as UAVs with interchangeable sensors and payload weight, travel time variability of the UAVs, limitation in fuel consumption, and time-dependent target visiting tasks. |
| Decentralized Critical Task Allocation Algorithm (DCTAA) | [57] | The authors employed DCTAA method with a three-phase iterative structure that in the first phase allocates tasks either by adding or forcing, in the second phase provides procedures for conflict resolution and in the third phase remove tasks from agents that overstepped their task capacity. |
| Phase Mission Analysis | [58] | The role of the mission analysis phase is to identify the likelihood of failure of the mission in each phase subject to completing all previous phases successfully. The approach is tailored based on carrying out a majority of the processing off-line in advance through the use of a library of probable failure causes of all potential phases. |
| Ontology-based Mission Planning Framework | [59] | The authors emphasized to move from a way-point representation of mission planning towards goal-based planning, while including multiple collaborating vehicles scenarios. Concept of SA-based adaptive autonomy is introduced as a more suitable approach rather than re-planning the mission for dealing with the environmental dynamism and vehicles capabilities. |

**Table.2.** Summary of recent approaches to AUV-based Mission Planning and Mission Management Systems

| Techniques | Ref | Highlights |
|---|---|---|
| Adaptive Track-Spacing Algorithm | [60] | The method applied on AUV optimal route planning approach in mine countermeasures mission in which the operation is framed in terms of maximizing underwater mines detection. No onboard processing capability is incorporated in the route planning resulting in non-adaptiveness of the approach to dynamic changes. |
| Ontology-based Architecture using a Hierarchical Analytic Process | [61] | The study modeled route planning with a multi-criteria decision-making mechanism. In the proposed strategy, the domain-specific ontology provides a foundation for maintaining or extending the domain knowledge. |
| Service-Oriented Architecture using Goal-based Planning Approach | [62] | This methodology enables domain-independent planning and capability of discovery for complex missions in which the system can handle poor and intermittent communications by predicting the intent of the other platforms to accomplish goal selection. The problem associated with knowledge-based (ontology-based) systems is that consistent information about the correct action and environmental situations is not always available. |
| Mixed Integer Linear Programming (MILP) | [63] | This approach applied MILP method for governing multiple AUVs. However, the MILP is a deterministic method, which is inaccurate in large sized problems as its computational time increases exponentially with the problem size. Hence, the real-time performance of the application is overshadowed by the growth of the graph complexity or problem search space. |
| Routing based on Dijkstra and GA | [64] | Performance of the Dijkstra and GA on vehicle routing problem is compared in this research. The results affirmed that both algorithms provided the same solution but Dijkstra's is more time-consuming compared to GA. |
| Route-task planning using metaheuristics of PSO, BBO and GA | [65] [66] | The large-scale AUV route-task planning problem has been addressed using heuristic search nature of PSO and GA on a static operation network in [65] and then expanded to a semi-dynamic uncertain network in [66], in which BBO and PSO employed to address AUV's task assignment and risk management joint problem. The merit of the proposed solution by [65,66] was independent of the graph size and complexity. |
| A three-layer hierarchical mission planning approach | [67] | The study is focused on development of a hierarchical mission planning approach including intuitive user interface, onboard mission re-planning software, and transit planning software to address environmental and risk-related factors such as time-varying subsurface currents, water levels, areas or times of high physical risk. |
| Hierarchical Petri Nets (PN) based Mission Management and Control System | [68] | The system allows the AUV to process multiple discrete events, where the highest level PN models the execution of the mission phases and the lower level sub-PNs provide mission control. The architecture is evaluated with lake trials in which promising results are achieved. |
| Behavior-based Hybrid Reactive/Deliberative Architecture | [69] | The proposed adaptive architecture uses both predictive and behavior-based approaches to handle under-informed situations and to control the deployment, activation, and deactivation of the behaviors to fulfill the mission. The feasibility of the architecture is assessed using AUV AVALON in a pipeline leak detection scenario. |
| PN –based Intelligent Decision Algorithm | [70] | The study considered environmental constraints such as forbidden zones, unsafe zones, and obstacle areas. A global path optimizer, and a speed optimization in combination with a hierarchical on-board architecture is utilized for executing the in-depth oceanic survey mission. This study is advantageous due to synchronous event handling, programmability, reactivity, and adaptability for the utilized mission management architecture. |
| Robust Mission Planning using deliberative Autonomy | [71] | The study dealt with abstracted diagnosis and failure recovery aiming to provide solutions for the robust, sustained presence of AUV in harsh underwater missions such as in-depth oceanic surveys. The study identified some critical issues in the design of AUV-based mission management system including: failure to turn up on time; uncertainty in onboard computation; sensor failure conditions; impaired mobility; and lack of built-in SA. |
| Modular Mission Control-Plan-Monitor-Replan Architecture | [72] | The study tackled the AUV's mission re-planning complexity in surveying in-depth oceanic scenarios in which the monitoring module is responsible for mission/plan initiation and observing mission execution while mission re-planning module is responsible for detecting the necessity of plan modification. The chart server module is tasked to check the modified plan against digital charts to confirm its consistency with terrain constraints. |
| Reactive-Deliberative Control Architecture | [73] | The approach provides a higher level of decision autonomy for an AUV to accomplish multiple tasks in a single mission in the face of periodic disturbances, turbulence, and uncertainties. The system incorporates two deliberative and reactive execution layers to satisfy autonomy requirements in both mission scheduling and motion planning for generating prompt actions/reactions. |
| Hybrid Synchronous Mission and Motion Planning System | [74] [75] [76] | These studies provided vehicle's decision autonomy through designing a hybrid synchronous system for ordering the execution of the tasks and guiding the AUV toward the target of interest in the higher layer; and using a SA-based motion planner as an action generator to ensure a safe and efficient deployment. They mostly focused on managing mission time and increasing mission productivity, while the effect of water current force as an important concern for AUV's deployment is not addressed. |
| Modular Deliberative-Reactive Control Architecture | [77] [78] | They contributed an autonomous interactive modular approach where the deliberative layer manages the concurrent execution of several tasks with different priorities and the reactive layer manages real-time reactions and performs a quick response to critical events. They emphasized a more realistic underwater environment comprising dynamic ocean current and uncertain moving/afloat objects, while the motion planner also is integrated with dynamic re-planning capability to deal with ocean dynamicity. The approach steps to reduce the reliance on operators that contribute to scalability, reliability and repeatability of operations. |

**Table.3.** Summary of recent approaches to Mission Planning and Mission Management Systems for Swarm Robotic Scenarios

| Techniques | Ref | Highlights |
|---|---|---|
| Route-Plan-Decision-Maker | [80] | The outlined study focused on dividing jobs between multiple robots while they are moving towards a destination point and managing the loose of any of robots by reassignment of its tasks into the closest groupmate. In this study, the environment is assumed to be ideal, while there are many uncertainties the in the real world. |
| Mission Planning based on GA and Ant Colony System (ACS) | [81] | In the study, GA is utilized for mission planning and identifying the minimum number of required UAVs to be engaged in the surveillance mission over the 3D environment. ACS is utilized for path planning to minimize the required changes in the traveling altitude of each UAV in the team while following a shortest path. |
| Temporary Command Control Method | [82] | They addressed multi-vehicle collaborative operation through the use of a temporary command control that allows any UAV in the team to take over the control of the team and ask other UAVs to readjust their position |
| | | in order to allow sufficient manoeuvring space/distance whenever a re-planning is essential. The approach keeps the human operators in the loop by allowing them to overrule any action if it is necessary. |
| Hierarchical Onboard Mission Planning Architecture | [83] | The onboard mission planning module works in a hierarchical manner that plans missions at the highest level (for multiple teams of UAVs) in mid-level generates plans for each team in the swarm and individual UAV at the lowest level. The generated plans consider resource allocation, payload configuration, collaborative use of gathered information, communication, target planning, role and responsibility allocation, task coordination, dynamic team forming, and interacting with external assets. |
| Interaction Force and Behavior Regulation Method | [84] | The study focused on multiple robots' coordination problem in congested systems and proposed amplification of the robots' velocities through the modified behavior regulation. The methodology is evaluated in simulations and is proved to be effective for the coordination of multiple robots in a congested system with bottlenecks. |
| RoboNetSim Simulator | [85] | They introduced a simulated environment called RoboNetSim capable of modeling network constraints for multi-robot scenarios, which provides network efficiency measurements for generated scenarios compatible with the real-world environments. |
| Swarm Controlling Approach Based on Crossing Pedestrian Flows (CPFs) Model | [86] | The study introduced a swarm controlling approach inspired by the appearance of congestion in CPFs, where the continuum model of CPF is able to generate density factor from dynamic changes of the congestion. The authors improved the average velocity by utilizing a control method and the feasibility of the approach is assessed through a simulation study. |
| Learning Automata-Based Probabilistic Algorithms and the Response Threshold Model (RTM) | [87] | They investigated heterogeneous task selection in multi-robot coordination based on generating decentralized solutions by allowing robots individually select tasks with the constraint of maintaining optimal task distribution. The simulation results indicated that the maximum principle is a better fit for the task selection, and the RTM is robust to noise and it is more time-intensive comparing to learning automata-based probabilistic approach. |
| Combination of High-Level Structures And a bottom-up approach | [88] | The study focused on navigating a swarm of robots, where the use of a bottom-up layer provides some degree of autonomy for individual robots, while the convergence in complex and critical circumstances is guaranteed via the use of a higher-level layer of intelligence. The advantage of approach in comparison with the top-down hybrid architectures is the ability to incorporate various algorithms designed for handling certain circumstances. |
| Smoothed-Particle Hydrodynamics (SPH) Simulation Technique | [89] | The approach forbids independent decision making within the swarm and utilizes a single algorithm to control the swarm, which is advantageous since the swarm is no longer dependent on individual members and the generated solution is robust concerning dynamic alternation of the swarm population. |
| Formation Control Law Utilizing Feedback Linearization with a Model Predictive Control (MPC) | [90] | They studied the formation of UVs' swarm and proposed a formation control law with the ability to convert the optimal control problem to a mixed integer quadratic programming problem, while obstacle avoidance is handled using Branch-and-Bound method. The numerical experiments showed significant computation time reduction. |
| System Identification Method based on Dependent Gaussian Process | [91] | The system is designed to handle modeling constraints of stability and control derivations of UAVs considering modeling constraints of dependency on model's prior knowledge and not being capable of addressing a wide flight envelope. The method is advantageous due to having efficient identification of coupling flight modes, a higher level of robustness to uncertain estimations, and being applicable to wider flight envelopes. |
| Area Extended PSO (AEPSO) and Cooperative AEPSO (CAEPSO) | [92] [93] | They investigated the potential of two PSO-based approaches on the problem of controlling a small swarm of simple robots in search and rescue scenarios, where the CAEPSO shows better ability to handle a combinatorial use of noise and learn from previous sessions. AEPSO performed a better local search, random search and linear search in both time-dependent and dynamic scenarios. The robustness of the AEPSO and CAEPSO to noise and task-time-dependency is considered to be the result of having better cooperation between agents achieved by knowledge sharing and balancing the use of essential behaviors of the swarm. |
| Integrated Mission Planning Strategy | [94] | The study developed an integrated mission planning strategy to serve the AUVs' routing and tasking problem, while minimizing total energy cost in the presence of ocean currents was the main objective of the study. |
| Behavior-based Controller Coupled with Waypoint Tracking Scheme | [95] | The study focused on double vehicle routing-task planning problem with multiple stacks, where the routing problem is modeled with a Double TSP with Multiple Stacks for a single-vehicle pickup-and-delivery problem. They studied the performance of three branch-and-cut, branch-and-price, and branch-and-price-and-cut algorithms on a wide family of benchmark test samples with different graph complexities. |
| Approximated Algorithm Based on a K-Means Method | [96] | The approach focused on addressing the task allocation problem that minimizes the travel time for multiple AUVs considering the presence of constant ocean current. The problem is considered in 2D that would be problematic when the method is applied to accomplish the similar task in a 3D environment. They also considered a static ocean current rather than the variable current, which is not prevalent in a real-world application. |
| Self-Organizing Map Neural Network (SOM-NN) | [97] | The study introduced SOM-NN with a velocity synthesis approach for solving multiple AUVs' dynamic task assignment and path planning problem. They considered multiple target locations in a 3D environment and dynamic current. The environment is assumed ideal without considering any kind of obstacle or uncertainty. |
| Strategy Over-Lord (SOL) System | [98] [99] | They introduced SOL system designed to oversee the moment-to-moment operations of multiple teams of heterogeneous UVs with the ability to detect near future events and proposing required changes to the robot's behaviors to the operators, which is capable of offering different solutions based on factors such as distance, associated direct and indirect risk of an enemy encounter, maneuverability of the proposed path, and network coverage. The system is capable of prioritizing between the proposed solutions and also allow the operators to interfere with the made decisions by SOL by rejecting the proposed actions. |